\documentclass{article}
\PassOptionsToPackage{numbers, compress}{natbib}
\usepackage[preprint]{neurips_2019}
\usepackage{multirow}
\usepackage[dvips]{graphicx}
\usepackage{amssymb, amsmath}

\usepackage{subfig}
\usepackage[utf8]{inputenc}
\usepackage{graphicx}

\usepackage{color}
\usepackage{units}

\usepackage{hyperref}

\setcitestyle{square}

\usepackage{float}
\restylefloat{table}
 
\begin{document}

\title{Adversarial Attacks on Deep Algorithmic Trading Policies
}

\author{Yaser Faghan, 
\thanks{Instituto Superior de Economia e Gest\~ao and CEMAPRE, Universidade de Lisboa, Portugal.
\tt{yaser.kord@yahoo.com}
}
\And
Nancirose Piazza,
\thanks{
Secure and Assured Intelligence Learning (SAIL) Lab, University of New Haven, New Haven, USA.
\tt{npiaz1@newhaven.edu}
}
\And
Vahid Behzadan,
\thanks{
Secure and Assured Intelligence Learning (SAIL) Lab, University of New Haven, New Haven, USA.
\tt{vbehzadan@newhaven.edu}
}
\And 
Ali Fathi
\thanks{
Enterprise Model Risk Management Group, Royal Bank of Canada (RBC).
\tt{ali.fathi@rbc.com}
}
\thanks{All contents and opinions expressed in this document are solely those of the author and do not represent the view of RBC Financial Group.}
}
\maketitle

\begin{abstract}
Deep Reinforcement Learning (DRL) has become an appealing solution to algorithmic trading such as high frequency trading of stocks and cyptocurrencies. However, DRL have been shown to be susceptible to adversarial attacks. It follows that algorithmic trading DRL agents may also be compromised by such adversarial techniques, leading to policy manipulation. In this paper, we develop a threat model for deep trading policies, and propose two attack techniques for manipulating the performance of such policies at test-time. Furthermore, we demonstrate the effectiveness of the proposed attacks against benchmark and real-world DQN trading agents.
\end{abstract}

\noindent\textit{Keywords and phrases}{
Deep Reinforcement Learning, Deep Q-Learning, AI Security, Capital Markets, Algorithmic Trading, Model Risk Management}

\section{Introduction}
The pursuit of intelligent agents for automated financial trading is a challenge that has captured the interest of researchers and analysts for decades \cite{Agbook}. The process of trading is well depicted as an online decision making problem involving two critical steps of \emph{summarizing the market condition} and \emph{execution of optimal actions}. For many years, algorithmic trading suffered from various problems ranging from difficulties in representations of the complex market conditions to real-time approaches to optimal decision-making in the trading environment. With recent advances in Machine Learning (ML), particularly in Deep Learning and Deep Reinforcement Learning (DRL), such challenges are dramatically alleviated via numerous novel proposals and architectures that enable end-to-end approaches to algorithmic trading \cite{deng2017deep}. In this context, end-to-end refers to the direct mapping of high-dimensional raw market and environment observations into optimal decisions in real-time. As data-driven agents, many such algorithms rely on sources of data that are either externally or collectively maintained, examples of which include market indicators \cite{Agbook} and social indicators (e.g., sentiment analysis from Twitter feeds \cite{cs229_2}).

While the growing interest in adoption of DRL techniques for algorithmic trading is well-justified by their impressive success in other domains, the risks involved in adversarial manipulation of such systems are yet to be explored. Recent developments in the domain of adversarial machine learning have brought attention to the security challenges in regards to the vulnerability of machine learning models to adversarial attacks \cite{SOK}. Instances of such attacks include adversarial examples \cite{fgsm}, which are strategically induced perturbations in the input vectors that are not easily detectable by human observers. Adversarial examples can be crafted through the calculation of an adversarial perturbation vector $\delta$ by solving the following iterative optimization problem: 
\begin{equation*}
\label{eq:estimated_reward_mean}
  {x^*} = \arg\min_{\delta_{x}}  \|\delta \| 
  \ {\text{ where }} \ 
   f(x+\delta) = t
\end{equation*}
Where $x$ is the correct classified example, $x^*$is the adversarial example, $f(.)$ is a classifier function, and $t$ is a class label other than the correct label $f(x)$. 

Adversarial attacks can impact all deep learning and classical machine learning models, including DRL agents \cite{Behzadan2018}. Recent work by Behzadan \cite{Behzadan, Behzadan2018, Behzadan2019} establish that DRL algorithms are vulnerable to adversarial actions at both training and inference phases of their deployment. This discovery is further verified in settings such as video games \cite{huang2017adversarial}, robotics \cite{clark2018malicious}, autonomous navigation \cite{behzadan2019adversarial}, and cybersecurity \cite{han2018reinforcement}. Yet, the extent, severity, and the dynamics of such vulnerabilities in DRL trading agents remain untouched. 

Adversarial perturbations of DRL trading policies are also significant form the financial Model Risk Managment (MRM) point of view (\cite{SR117}, \cite{E23}, \cite{modelRisk}) since the existence of such vulnerabilities can be traced back to the algorithmic underpinnings of these systems. However, principal differences between traditional financial models and algorithmic trading systems pose additional challenges for quantifying and containing the resulting model risk. For instance, the number of model components involved in an algorithmic trading system can be quite large and hence, fusion of otherwise individually negligible residual model risk may result in significant system errors. Furthermore, There exist the adaptive nature of DRL based algorithms where the model components are re-calibrated (e.g., through retraining) based on a low latency schedule. It should also be noted that unlike other areas of quantitative modelling in finance (such as asset pricing or credit risk) the benchmarking of various model components of Algorithmic systems are not possible due to competition considerations, as there may be restrictions for conducting open box validation of proprietary models within a firm.

 In this paper, we investigate adversarial attacks against DRL trading agents at test-time. Accordingly, the main contributions of this paper are:
\begin{itemize}
    \item We provide a DRL threat model for DRL trading agents, identifying susceptible attack surfaces and practical attack vectors at test-time.
    \item We demonstrate of feasibility of the proposed attack vectors in manipulating DRL trading agents at various level of complexity.
    \item We establish the vulnerability of current DRL trading policies to adversarial manipulation, and bring attention to the need for further analysis and mitigation of such attacks in deployed DRL trading agents.
\end{itemize}

\noindent The remainder of the paper is as followed: Section \ref{sec:Background} presents an overview of reinforcement learning and Deep Q-Networks (DQN), as well as a review of the security issues in electronic trading platforms. Section \ref{Sec:ThreatModel} proposes a DRL threat model for trading DRL agents, and outlines various attack surfaces and vectors that an adversary may leverage to manipulate the trading policies. Section \ref{sec:ExperimentDesign} provides the details of our experimental setup for investigating the proposed attack mechanism, the results of which are presented in Section \ref{sec:Attacks}. The paper concludes in Section \ref{sec:Conclusion} with a summary of our findings, as well as discussions on future directions of research on the security of deep trading policies.

\section{Background}
\label{sec:Background}
\subsection{Reinforcement Learning}
Reinforcement learning is concerned with agents that interact with an environment and exploit their experiences to optimize a decision-making policy. The generic RL problem can be formally modeled as learning to control a Markov Decision Process (MDP), described by the tuple $MDP = (\mathbb{S}, \mathbb{A}, \mathbb{R}, \mathbb{P})$, where $\mathbb{S}$ is the set of reachable states in the process, $\mathbb{A}$ is the set of available actions, $\mathbb{R}$ is the mapping of transitions to the immediate reward, and $\mathbb{P}$ represents the transition probabilities (i.e., state dynamics), which are initially unknown to RL agents. At any given time-step $t$, the MDP is at a state $s_t\in \mathbb{S}$. The RL agent's choice of action at time $t$, $a_t \in \mathbb{A}$ causes a transition from $s_t$ to a state $s_{t+1}$ according to the transition probability $\mathbb{P}(s_{t+1}|s_t , a_t)$. The agent receives a reward $r_{t+1} = \mathbb{R}(s_t, a_t, s_{t+1})$ for choosing the action $a_t$ at state $s_t$. Interactions of the agent with MDP are determined by the policy $\pi$. When such interactions are deterministic, the policy $\pi: \mathbb{S}\rightarrow \mathbb{A}$ is a mapping between the states and their corresponding actions. A stochastic policy $\pi(s)$ represents the probability distribution of implementing any action $a\in \mathbb{A}$ at state $s$. The goal of RL is to learn a policy that maximizes the expected discounted return $E[R_t]$, where $R_t = \sum_{k=0}^{\infty}\gamma^k r_{t+k}$; with $r_t$ denoting the instantaneous reward received at time $t$, and $\gamma$ is a discount factor $\gamma\in [0,1]$. The value of a state $s_t$ is defined as the expected discounted return from $s_t$ following a policy $\pi$, that is, $V^{\pi}(s_t) = E[R_t|s_t, \pi]$. The action-value (Q-value) $Q^{\pi}(s_t, a_t) = E[R_t|s_t,a_t, \pi]$ is the value of state $s_t$ after applying action $a_t$ and following a policy $\pi$ thereafter. 


\subsection{Value Iteration and Deep Q-Network}
Value iteration refers to a class of algorithms for RL that optimize a value function (e.g., $V(.)$ or $Q(.,.)$) to extract the optimal policy from it. As an instance of value iteration algorithms, \emph{Q-Learning} aims to maximize for the action-value function $Q$ through the iterative formulation of Eq.~(\ref{bellman}):
\begin{eqnarray} \label{bellman}
Q(s,a) = R(s, a) + \gamma max_{a'}(Q(s',a'))
\end{eqnarray}

Where $s'$ is the state that emerges as a result of action $a$, and $a'$ is a possible action in state $s'$. The optimal $Q$ value given a policy $\pi$ is hence defined as: $Q^{*} (s, a) = max_{\pi} Q^{\pi} (s, a)$, and the optimal policy is given by $\pi^{*}(s) = \arg\max_a Q(s,a)$.

The Q-learning method estimates the optimal action policies by using the Bellman formulation to iteratively reduce the \emph{TD-Error} given by $Q_{i+1} (s,a) - \mathbf{E}[r + \gamma \max_a Q_i]$ for the iterative update of a value iteration technique. Practical implementation of Q-learning is commonly based on function approximation of the parametrized Q-function $Q(s,a; \theta) \approx Q^\ast (s,a)$. A common technique for approximating the parametrized non-linear Q-function is via neural network models whose weights correspond to the parameter vector $\theta$. Such neural networks, commonly referred to as Q-networks, are trained such that at every iteration $i$, the following loss function is minimized:
\begin{eqnarray}\label{eq:update}
L_i(\theta_i) = \mathbf{E}_{s, a\sim \rho(.)} [(y_i - Q(s,a,;\theta_i))^2]
\end{eqnarray}

where $y_i = \mathbf{E}[r + \gamma \max_{a'}Q(s',a';\theta_{i-1}) | s,a]$, and $\rho(s,a)$ is a probability distribution over states $s$ and actions $a$. This optimization problem is typically solved using computationally efficient techniques such as Stochastic Gradient Descent (SGD).

Classical Q-networks introduce a number of major problems in the Q-learning process. First, the sequential processing of consecutive observations breaks the \emph{i.i.d.} (Independent and Identically Distributed) assumption on the training data, as successive samples are correlated. Furthermore, slight changes to Q-values leads to rapid changes in the policy estimated by Q-network, thus giving rise to policy oscillations. Also, since the values of rewards and Qs are unbounded, the gradients of Q-networks may become sufficiently large to render the backpropagation process unstable.

A Deep Q-Network (DQN) \cite{Silver} is a training algorithm designed to resolve these problems. To overcome the issue of correlation between consecutive observations, DQN employs a technique called \emph{experience replay}: instead of training on successive observations, experience replay samples a random batch of previous observations stored in the replay memory to train on. As a result, the correlation between successive training samples is broken and the \emph{i.i.d.} setting is re-established. In order to avoid oscillations, DQN fixes the parameters of a network $\hat{Q}$, which represents the optimization target $y_i$. These parameters are then updated at regular intervals by adopting the current weights of the Q-network. The issue of instability in backpropagation is also solved in DQN by normalizing the reward values to the range $[-1,+1]$, thus preventing Q-values from becoming too large.

Mnih et al. \cite{Silver} demonstrate the application of this new Q-network technique to end-to-end learning of Q values in playing Atari games based on observations of pixel values in the game environtment. To capture the movements in the game environment, Mnih et al. use stacks of four consecutive image frames as the input to the network. To train the network, a random batch is sampled from the previous observation tuples $<s_t, a_t, r_t, s_{t+1}>$, where $r_t$ denotes the reward at time $t$. Each observation is then processed by two layers of convolutional neural networks to learn the features of input images, which are then employed by feed-forward layers to approximate the Q-function. The target network $\hat{Q}$, with parameters $\theta^{-}$, is synchronized with the parameters of the original $Q$ network at fixed periods intervals. i.e., at every $i$th iteration,  $\theta^-_{t} = \theta_t$, and is kept fixed until the next synchronization. The target value for optimization of DQN thus becomes:

\begin{eqnarray}\label{eq:DQNtarget}
y_t \equiv r_{t+1} + \gamma max_{a'} \hat{Q}(s_{t+1}, a'; \theta^-)
\end{eqnarray}

Accordingly, the update rule for the parameters in the DQN training process can be stated as:

\begin{eqnarray}\label{SGD}
\theta_{t+1} = \theta_t + \alpha(y_t - Q(s_t, a_t; \theta_t))\nabla_{\theta_t}Q(s_t, a_t; \theta_t)
\end{eqnarray}

\subsection{State of Security in Algorithmic Trading}
In recent years, electronic trading platforms have made access to global capital markets easier for the general public, resulting in a lower barrier to entry and an influx of traffic across these platforms. The growing interest in such trading platforms and technologies is however accompanied by the increasing risks of cyber attacks. While the literature on the cybersecurity issues of current trading platforms is scarce, few industry-sponsored studies report concerning issues in deployed trading platforms. One such study on the exposure of security flaws in trading technologies \cite{Hernandez2018} evaluates various popular desktop, mobile and web trading service platforms against a standard list of security checks, and reports that these trading technologies are in general far more susceptible to cyber attacks than previously-reviewed personal banking applications from 2013 and 2015. The security checks consisted of features such as 2-Factor Authentication (2FA), encrypted communications, privacy mode, anti-reverse engineering, and hard-coded secrets. This study reports that 64\% of the reviewed trading applications rely on unencrypted communication channels for authentication and trading data. Also, the author finds that many trading applications utilize poor session management and SSL certificate validation, thereby enabling Man-in-The-Middle (MITM) attacks. Furthermore, this report points out the wide-scale susceptibility of such platforms to remote Denial of Service (DoS) attacks, which may render the applications useless. Building on the findings of this study, our paper investigates attacks that leverage the aforementioned vulnerabilities to manipulate deep algorithmic trading policies. 

\section{Threat Model of DRL Trading Agents}
\label{Sec:ThreatModel}
Adversarial attacks against DRL policies aim to compromise one or more aspects of the Confidentiality, Integrity, and Availability (CIA) triad in the targeted agents \cite{Behzadan2018}. More specifically, the \emph{Confidentiality} of a DRL agent refers to the need for confidentiality of an agent's parameters, such as the policy or reward function. The \emph{Integrity} of a DRL agent relies on the policy behaving as intended by the user. \emph{Availability} refers to the agent's capability to execute its task when needed. At a high-level, the threat landscape of DRL agents can be captured in terms of the \emph{Attack Surface} and \emph{Attack Model} of the agent \cite{Behzadan2019}, as outlined below.

\subsection{Attack Surface and Vectors}
Adversarial attacks may target all components of a DRL agent, including the environment, agent's observation channel, reward channel, actuators, and training components (e.g., experience storage and selection), as identified by Behzadan \cite{Behzadan2019}. 

\begin{figure}[htbp]%
    \centering
{{\includegraphics[width=\columnwidth]{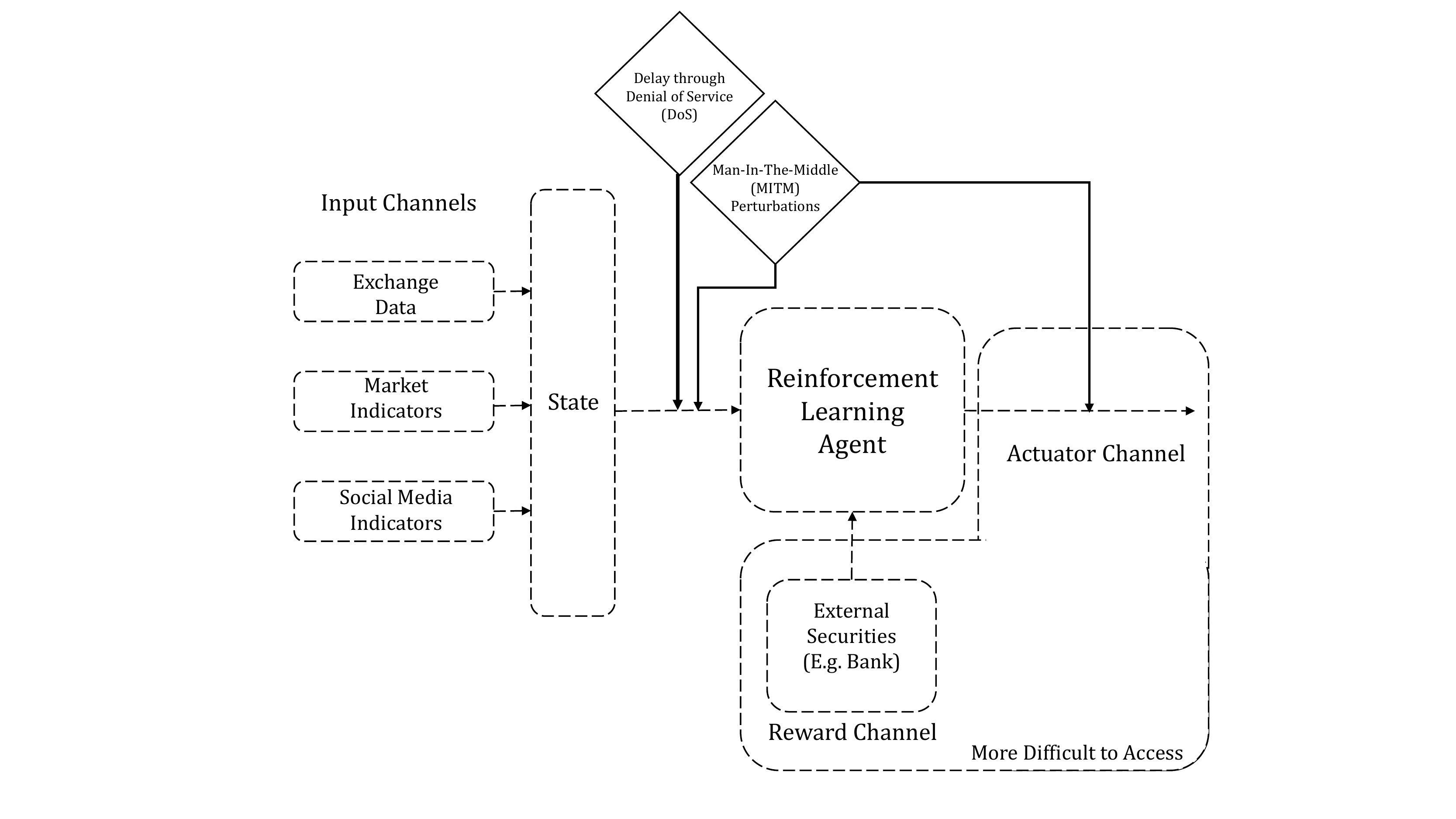} }}%
    \caption{Attack Surface and Vectors of a DRL Trading Agent at Test-Time}%
    \label{diagram}%
\end{figure}

Figure \ref{diagram} illustrates the components of a DRL trading agent at test-time. In the context of algorithmic trading, the observation of the environment is gathered from various sources such as market indicators, social media indicators, and exchanges-- we refer to these sources as input channels. This data is prepossessed and feature engineered to create the agent's observation of the state. These states are part of the observation returned by the environment to the agent along with the reward. Through the observation channel, an adversary may intercept the observation and exchange it for a perturbed observation, otherwise called a Man-In-The-Middle (MITM) attack. An adversary may also delay the observation channel through a Denial of Service (DoS) attack. The reward channel is often tied to internal securities such as bank accounts or portfolios, and hence are less susceptible to external adversarial manipulation. However, any external component reachable by the agent can be compromised implicitly. 

\subsection{Attack Model}
The capabilities of an adversary are defined by two factors, namely the actions available to the adversary, and information available about the target. These define the extend and impact of eligible attacks on a DRL agent. This section presents a classification of attacks and adversaries at the inference phase based on the aforementioned factors. 


Inference-time (also referred to as test-time) attacks may be active or passive. Passive attacks aim to gather information about the target agent by observing the behavior of the target in various states. With sufficient observations of state-action pairs, the adversary can reconstruct the targeted policy and thereby compromise the Confidentiality of the targeted, proprietary agents \cite{BehzadanImitation}. On the other hand, active attacks are those that perturb one or more components of a DRL agent to manipulate the behavior of its policy. According to the available information, such attacks can be classified as whitebox or blackbox. Whitebox attacks refers to cases where the adversary has sufficient knowledge of the target's parameter to directly craft an effective perturbation, and blackbox attacks refer to the vice versa scenario. Imitation Learning and Inverse Reinforcement Learning are avenues an adversary may use to either attack their target agent or steal components of the agent like its learned policy. As demonstrated in \cite{BehzadanImitation}, adversaries can gather additional information through policy imitation, thereby enabling whitebox attacks against blackbox targets.

Furthermore, active manipulations can be classified under targeted and non-targeted attacks. 
Successful non-targeted or indiscriminate attacks through aim to have the policy select any action other than the optimal action $a$ by providing a perturbed observation instead of the true observation at a timestep $t$, resulting in less reward given by the environment. Targeted attacks aim to craft perturbations such that the policy selects a particular sub-optimal action $a'$, as chosen by the adversary. For each attack, a perturbation vector is crafted that is minutely different to the true observation vector that would otherwise be undetectable by human traders, but produces state values that result in the policy choosing a different action than action $a$. This is similar to how adversarial attacks affect supervised machine learning models. 

Perturbations in observation affect both test-time and train-time. While the focus of this paper is on test-time attacks, it is noteworthy that during training, perturbed observations result in state values which are bootstrapped throughout its updates, potentially impacting learned policies and their trajectory which is the sequence of actions taken by the agent. Work by Behzadan and Munir \cite{behzadan1} show that training-time attacks under certain conditions with high probability of attack per observation between 0.5 and 1.0 resulted in the agent's inability to recover performance upon test-time evaluation under non-adversarial conditions.



\section{Experimental Setup}
\label{sec:ExperimentDesign}

We consider two DQN agents and their environments to demonstrate the proposed attacks, one we will refer to as the basic DQN which uses a simple OpenAI Gym\footnote{ OpenAI Gym, (2016), GitHub repository, https://github.com/openai/gym} environment to emulate trading, and the other is based on an open-source framework called TensorTrade\footnote{ TensorTrade, (2019), GitHub repository, https://github.com/tensortrade-org/tensortrade} which leverages a more realistic OpenAI Gym environment mimicking real-world trading settings. 
Our basic DQN represents less complex agents while TensorTrade's DQN will demonstrate the real-world impact of such attacks that have external components tied to the agent like a portfolio. In fact, TensorTrade is currently used and deployed for actual DRL-based trading in online cryptocurrency and stock exchanges.

\subsection{Basic Trading Environment} 
In this scenario, our data is sourced from Russian stock market prices between the period of 2015-2016. The dataset is comprised of samples representing a one-minute temporal resolution, and the dynamic of the price during that minute is captured by four values:
\begin{itemize}
  \item open price -  price at the beginning of the minute
  \item high price -  maximum price during the minute interval
  \item low price -  low is the minimum price during the minute interval
  \item close price - last price of the minute time interval
\end{itemize}
Each minute interval is called a bar.
Our agent will have three actions: Buy a Share, Wait, or Close the Position/Sell. Hypothetically, to buy a share, the agent would borrow a stock share and be charged a commission. If the agent already owns a share, nothing will happen and the agent can have at most one share. If an agent owns a share and decides to close the position/sell, the agent will pay a commission which is a fixed percentage of the current price. If the agent does not own a share, nothing will happen. The action \emph{Wait} is that of taking no action at all. 
Table~\ref{table:1} details the specifications of the Basic Stock Environment. Table~\ref{table:2} contains hyperparameters of the DQN agent trained in this environment.

\begin{table}[htbp]
\caption{Specifications of the Basic Stock Environment}
\label{table:1}

\begin{center}
\begin{tabular}{|l|l|l|}
\hline

\multirow{6}{5em}{Observation Space}
&& - Past 10 bars/tuples (relative to open price):\\
&& ---- relative high price\\
&& ---- relative low price\\
&& ---- relative close price\\
 && - Indication of bought share [0 or 1] within the window of 10 past bars \\
 && - Profit or loss from current position\\
 \hline
\multirow{3}{5em}{Action Space} && - Buy a Share \\
&  & - Wait \\
 &  & - Close the Position (Sell) \\ \hline
\multirow{3}{5em}{Reward}  && - Without Position: [100 * ( SP-BP ) / BP ]\% - C\% \\ 
&& - With Position: - C\% \\
&& SP is Sold Price, BP is Bought Price, C is Commission\\\hline

\multirow{1}{5em}{Termination} && Episode length is greater than 250 \\
\hline
\end{tabular}
\end{center}
\end{table}

\begin{table}[htbp]
\caption{Basic DQN Training Hyperparameters}  
\label{table:2}
\begin{center}
\begin{tabular}{|l|l|l|}
\hline
No. Time steps && $10^{5}$ \\ 
 \hline
$\gamma$ && $0.99$ \\
 \hline
Learning Rate && $10^{-4}$ \\
 \hline
Replay Buffer Size && $10^{5}$ \\
 \hline
First Learning Step && $1000$ \\
 \hline
Target Network Update Freq. && $1000$ \\
 \hline
Exploration && Parameter-Space Noise \\
\hline
Exploration Fraction && $0.1$ \\
\hline
Final Exploration Prob. && $0.02$ \\
\hline
Max. Total Reward && $250$ \\
\hline
\end{tabular}
\end{center}
\end{table}

\subsection{TensorTrade Environment} 

The TensorTrade environment (TT) has more components than the basic DQN's environment. TT can implement a portfolio that holds wallets of various coins or currencies. The data used for this setup is included with TT as a demonstration of training. This dataset is dated from the start of 2020, and contains the open, high, low, close and volume prices at hourly intervals. It also includes technical indicators such as the Relative Strength Indicator (RSI) and Moving Average Convergence Divergence (MACD) as well as $log(C_t) - log(C_{t-1})$ where $C_t$ is the closing price at timestep $t$ as the dataset features. Our portfolio starts with the same setup as TT's demo which includes 10,000 USD and 10 BTC. We use the risk-adjusted reward scheme and manage-risk action scheme provided by TT. The risk-adjusted reward scheme uses the Sharpe Ratio which is defined by the equation below:
$$S_a = \frac{E[R_a - R_b]}{\sigma_a} $$ where $R_a$ is the asset return, $R_b$ is the risk-free return, and $\sigma_a$ is the standard deviation of the asset excess return. The implementation offsets the numerator and denominator by a small decimal to avoid zero division. The manage-risk action scheme scales the action space depending on provided arguments such as trade size, stop and take. The default trade size is 10 which implies there will be a list of 10 trade sizes that are uniformly spaced. For instance, trade size of 3 implies $33.3\%$, $66.6\%$, and $99.9\%$ of the balance can be traded. \emph{Take} is a list of possible take profit percentages from an order, and \emph{stop} is a list of possible stop loss percentages from an order. The action space is the resulting product of take, stop, trade size, and action type which is buy or sell. There is one additional action, namely wait/hold placed at index 0. In our case, we have an action space size of 181. This information as well as training hyperparameters are summarized in Table \ref{tt_env} and Table \ref{tt_hp}, respectively. There are other simpler reward (e.g., SimpleProfit) and action (e.g., BSH stands for Buy Sell Hold) schemes available with TT. 

\begin{table}[htbp]
\caption{Specifications of TensorTrade Environment}
\label{table:1}

\begin{center}
\begin{tabular}{|l|l|l|}
\hline

\multirow{4}{5em}{Observation Space}
&& Past 20 feature vector tuples of:\\
 & & - log(C$_t$) - log(C$_{t-1}$) where C$_t$ is the closing price at timestep $t$ \\
 && - MACD (fast=10, slow=50, signal=5)\\
 && - RSI (period=20)\\
 \hline
\multirow{3}{5em}{Action Space} && Managed Risk Sceheme:\\
&&- Product(stop list, take list, trade size list, [buy, sell]) = 180 actions \\
&  & - Wait/hold action (indexed at 0) \\
 \hline
Reward  && Risk-Adjusted Scheme using Sharpe Ratio\\
\hline
Termination && Timestep $>$ 250 \\
\hline
\end{tabular}
\end{center}
\label{tt_env}
\end{table}

\begin{table}[htbp]
\caption{TensorTrade's DQN Training Hyperparameters}  
\label{table:2}
\begin{center}
\begin{tabular}{|l|l|l|}
\hline
No. Timesteps && 250 \\ 
 \hline
 Episodes && 100 \\ 
 \hline
 Epochs && 80 \\ 
 \hline
$\gamma$ && $0.9999$ \\
 \hline
Learning Rate && $10^{-5}$ \\
 \hline
Replay Buffer Size && $10^{3}$ \\
 \hline
Target Network Update Freq. && $10^{3}$ \\
 \hline
Exploration && $\epsilon$-greedy \\
\hline
Optimistic Initialization $\epsilon$ && $0.9$ \\
\hline
Minimum $\epsilon$ && $0.05$ \\
\hline
Decay $\epsilon$ every N steps && 200 \\
\hline
\end{tabular}
\end{center}
\label{tt_hp}
\end{table}

\section{Attacks}
\label{sec:Attacks}
In this section, we investigate the impact of adversarial attacks on deep trading agents at test-time. To preserve the realism of our study, we limit the scope of our investigation to attacks that satisfy the following constraints:
\begin{enumerate}
    \item Attacks are limited to manipulating the observation channel of the target.
    \item Attacks are limited to perturbations that are not immediately detected by common human or automated anomaly detection mechanisms.
\end{enumerate}

We consider two modes of attacks on the observation channel of the DRL trading agent: targeted, and non-targeted. Furthermore, we implement 2 different types of attack namely delay attacks, and adversarial example (i.e., perturbation) attacks. This study considers whitebox attacks only, implying that the adversary is assumed to have complete knowledge of the target's architecture and parameters. However, as demonstrated in \cite{BehzadanImitation}, it is also feasible to reverse-engineer blackbox policies via imitation learning, thereby converting blackbox attacks to whitebox. 

\subsection{Non-Targeted Delay Attacks}
\begin{figure}[htbp]%
    \centering
    \begin{minipage}{\textwidth}
    \hspace*{-2cm} 
    \subfloat[\centering Basic DQN]{{\includegraphics[width=0.5\columnwidth]{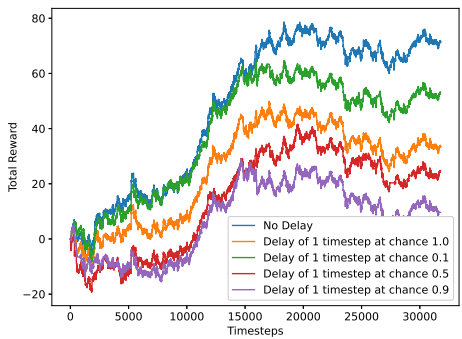}  }}%
    \qquad
    \subfloat[\centering TensorTrade DQN ]{{\includegraphics[width=0.7\columnwidth]{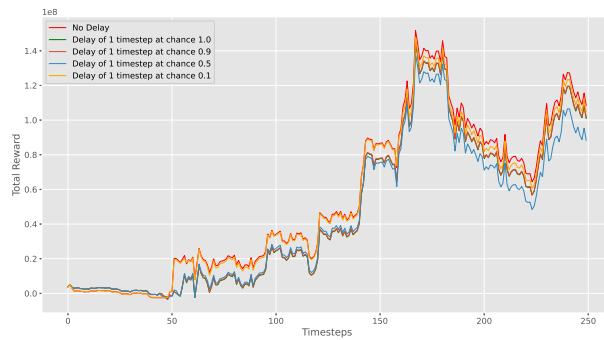} }}%
    \caption{Observational Delay}%
    \label{obsdelay}%
    \end{minipage}
\end{figure}

We evaluate through non-targeted attacks on a single, most recent window history tuple of their features used for observation. This is an attack on the observation channel. For the basic DQN, this would be a tuple of relative high, relative low and relative close prices in regards to their open price. For TensorTrade's DQN, this would be $log(C_t) - log(C_{t-1})$ where $C_t$ is the closing price at timestep $t$, MACD, and RSI. Our first attack will be an observation delay of 1 timestep where a tuple of values seen at timestep $t-1$ will be received at timestep $t$. We choose a delay of 1 timestep because it is both practical and representative of minimal interference. Results are presented in Figure~\ref{obsdelay}. As is demonstrate in the results, when an agent picks an action based on delayed observation from timestep $t-1$, action received by the environment which is at a true timestep $t$ returns a reward which can be at most equivalent to the optimal action reward at the timestep $t$. This type of non-targeted attack should be of concern to traders because of its lack of computational expense to implement, and adversarial predisposition since it depends on time-series locality to mask anomalies.

\subsection{Non-Targeted Perturbation Attacks}
To investigate the effectiveness of adversarial example attacks on DRL policies, we implemented Fast Gradient Sign \cite{fgsm} and Carlini and Wagner ( C\&W) \cite{cw} adversarial example attacks using $L_2$ loss for both DQNs. 

\begin{figure}[htbp]%
    \centering
    \begin{minipage}{\textwidth}
    \hspace*{-2cm} 
    \subfloat[\centering Non-Targeted FGSM Attack]{{\includegraphics[width=0.6\columnwidth]{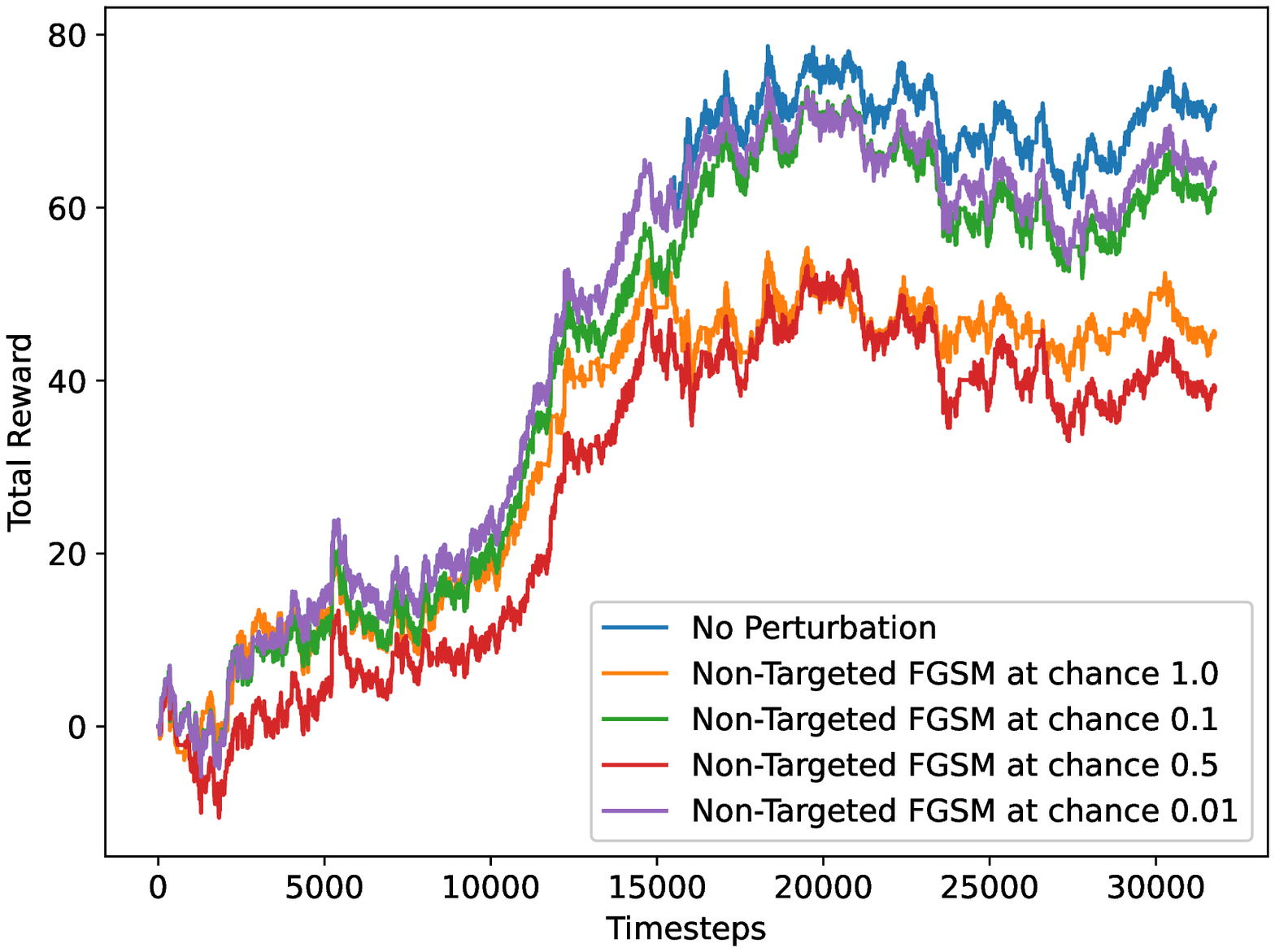}  }}%
    \qquad
    \subfloat[\centering Non-Targeted C\&W Attack ]{{\includegraphics[width=0.6\columnwidth]{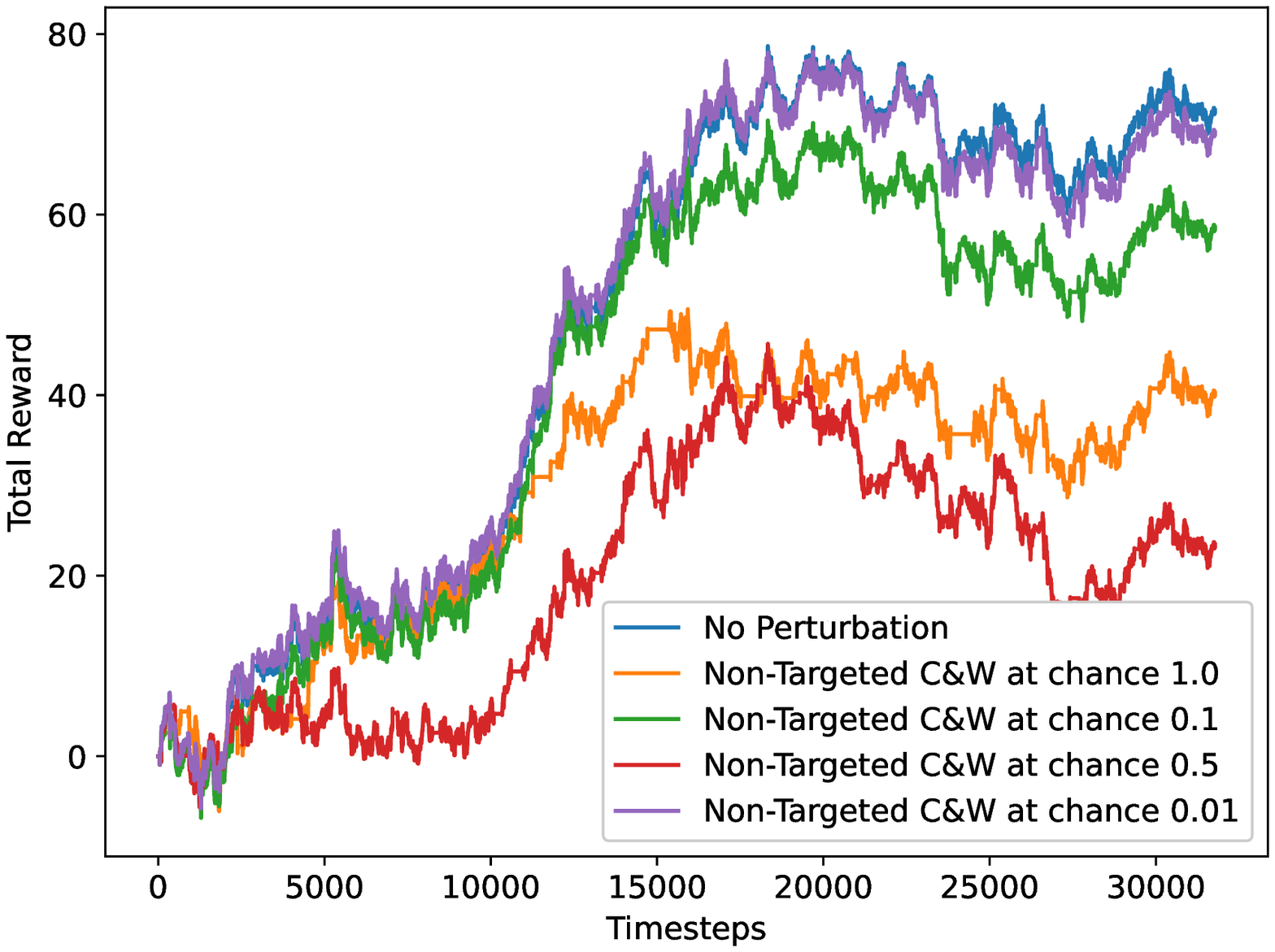} }}%
    \caption{Non-Targeted Attacks on the Basic DQN}%
    \label{non_target_attacks}%
    \end{minipage}
\end{figure}

For our basic DQN implementation, the values of high, low, close prices in the observation space are relative prices scaled according to their open price. The data is not normalized along its dimension, but the value range is similar by density since majority of the relative prices fall within a bounded region of values and its range is fairly close to $[0,1]$. Instead of selecting the perturbation threshold $\epsilon$ through data prepossessing such that $\epsilon$ can be a valid representation of allowable perturbation along a dimension, we test a range of small epsilon values and chose the best among the selection that produced the most adversarial-like samples. Additionally, we apply post constraints with a strong assumption of a fixed, uniformal step-size across all dimensions to be acceptable. We have fixed the initial $\epsilon$ to $0.0001$ and for any failure to pick another action at a timestep, the attack will be allowed five iterations at increasing $\epsilon$ values where its last iteration tests at an $\epsilon$ of value $0.001$. Our implementation of the C\&W attack on the basic DQN has a max iteration of 100 with a learning rate of 0.5 and initial constant of 0.1. We chose a lower number of iterations and higher learning rate because of computational expense, but the results are sufficient in representing adversarial observations that are similar to their true observation state. Results of the non-target FGSM attack and non-target C\&W attack on the basic DQN can be seen in Figure \ref{non_target_attacks}. Only successful non-target attacks have their perturbed observation saved for future timesteps. Additionally, if there is a successful attack at timestep $t$ that has impacted a future action a$_{t+k}$ where $k<0 \leq N$, we do not attack the timestep and count the timestep as no change needed (NCN). We define a failure for the basic DQN as the failure to change an agent's action. Our attacks abide by constraints that makes these perturbations adversarial when compared to the true observations: Relative high prices are non-negative and relative low prices are non-positive. Relative closing price must be bounded within the relative high price and relative low price. We also have considered matching the true relative close price's behavior whether it was a relative high price, relative low price, or strictly between the prices. By applying these post constraints, we shift the perturbation of a dimension within the correct distribution of values for a given dimension. Table~\ref{nontarget_table} contains failure counts and other notable counts for the basic DQN. Representative samples of a perturbed tuple of values from successful attacks are presented in Table \ref{fgsm_nontarget_samples} and Table \ref{cw_nontarget_samples}. 

Our TensorTrade DQN's action space is larger and we have fewer timesteps to attack, which imply a higher failure ratio for attacks with perturbation per observation probabilities less than 1.0. We adjust our definition of failure as the failure to change an agent's action $a$ to some action $a'$ which is not the same action type. For instance, if $a$ is a buy action type, then the attack has failed to change the agent's action to either a sell action type or wait action type. We have tighter constraints for non-targeted attacks because we are more interested in change of action type in regards to indiscriminate attacks. However, more lenient failure constraints are also valid since they also impact the agent's performance. Results on failure count and attempts can for TT's DQN can be found in Table \ref{tt_nontarget_table}. Unlike the basic DQN's observations, our observations are not bounded to a small range close to $[0,1]$ nor have similar distributions along its dimensions such that uniform steps in all dimensions are adversarial. This stems from the non-normalized state of the data. Additionally, normalizing data that is not innately bounded makes it difficult to guarantee that FGSM's $\epsilon$ which is used to represent the amount of allowable perturbation in all dimensions be applicable to newer data that may fall out of the existing normalized boundaries. Therefore, we cannot think of FGSM's $\epsilon$ as $\epsilon \in (0,1]$ where it represents allowable perturbation in this case. We evaluate values of $\epsilon$ by comparing the adversarial observation with its true observation. Since the features do not share similar ranges of values, we individually scale each perturbed feature by $\epsilon\times k_d$, where $0 \leq k_d\leq 1$, where $d$ is the respected feature/dimension such that our perturbations are similar to their true observation tuple. In our implementation, these $k$ scalars are exponential forms $10^{-m}$ where $m \in \mathbb{N}$. Alternative $k$ scalars can be calculated for each dimension by using their distribution. Our non-targeted FGSM attack follows the same setup as before for the basic DQN but we start at an $\epsilon$ of 0.1 and end at an epsilon of 3.0 with $k_{0} = 0.01$, $k_{1} = 0.01$, and $k_{2} = 0.1$. For example, if given a tuple:
$$ (x_0,x_1,x_2)$$
where these values are from our feature vector, then the perturbation tuple where $\delta = (p_0,p_1,p_2)$  will be $$(x_0+\epsilon k_0p_0,  x_1+\epsilon k_1p_1, x_2+\epsilon k_2p_2)$$.

\begin{figure}[htbp]%
    \centering
    \begin{minipage}{\textwidth}
    \hspace*{-2cm} 
    \subfloat[\centering Non-Targeted FGSM Attack Reward Difference]{{\includegraphics[width=0.6\columnwidth]{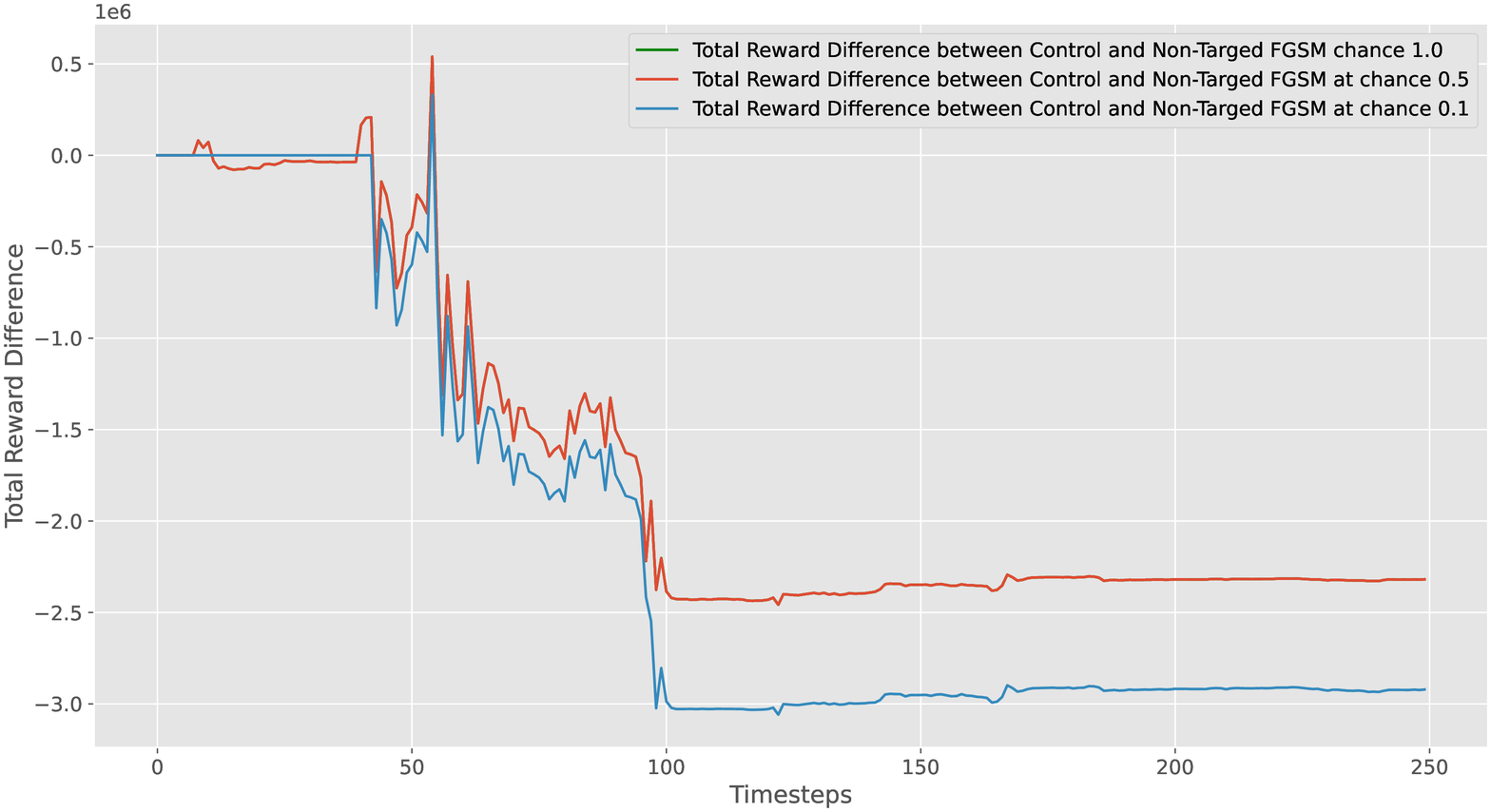}  }}%
    \qquad
    \subfloat[\centering Non-Targeted C\&W Attack Reward Difference ]{{\includegraphics[width=0.6\columnwidth]{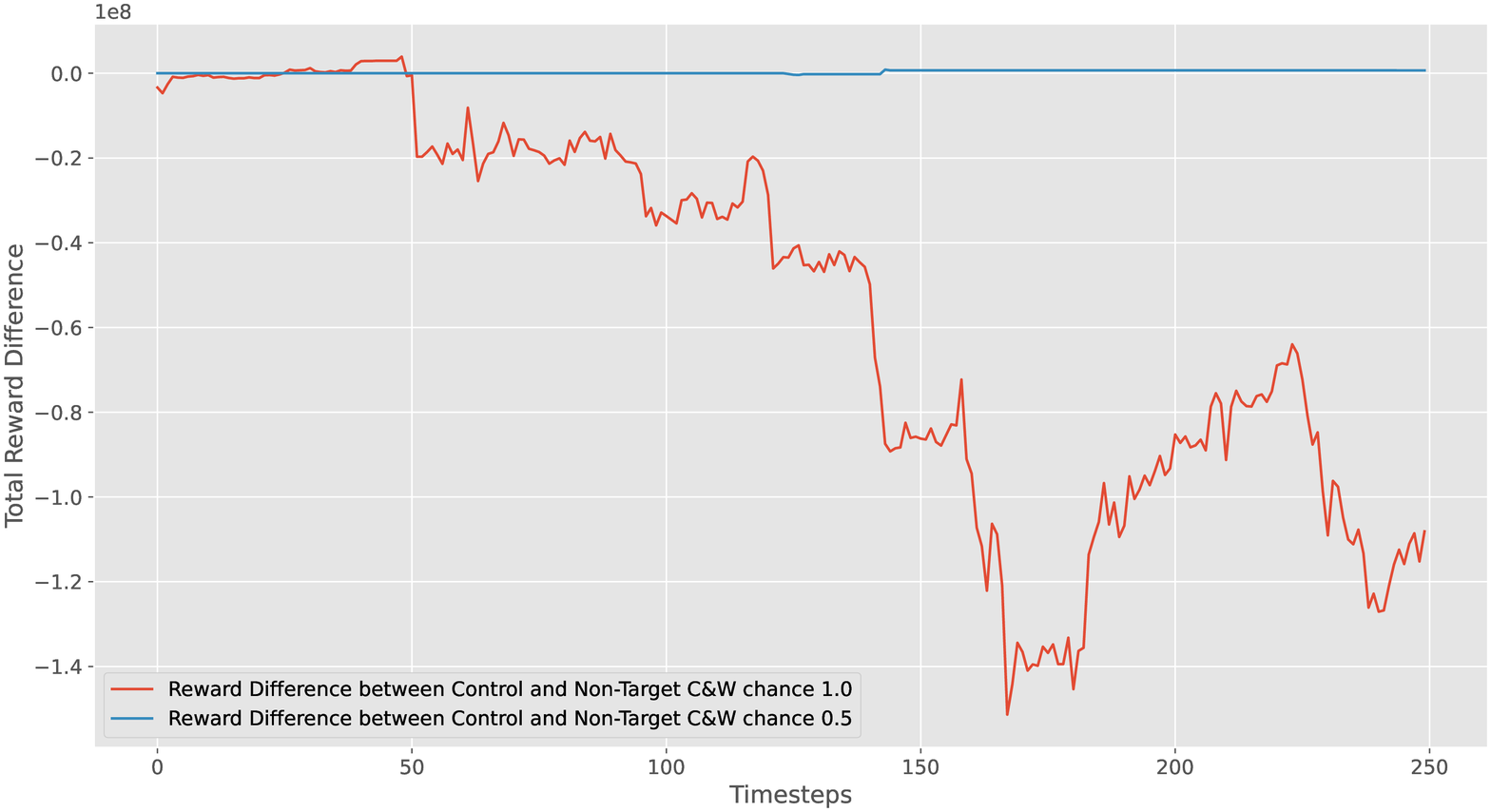} }}%
    \caption{Reward Differences between Control Total Reward and Non-Targeted Attacks on TensorTrade's DQN Total Reward}%
    \label{non_target_attacks_tt_rd}%
    \end{minipage}
\end{figure}

\begin{figure}[htbp]%
    \centering
    \begin{minipage}{\textwidth}
    \hspace*{-2cm} 
    \subfloat[\centering Non-Targeted FGSM Attack Total Net-Worth Difference]{{\includegraphics[width=0.6\columnwidth]{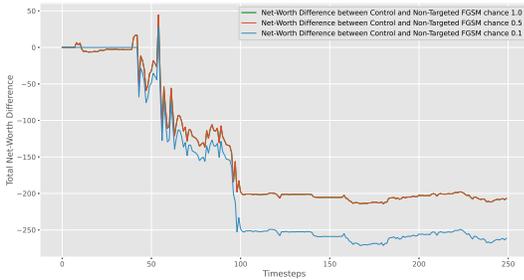}  }}%
    \qquad
    \subfloat[\centering Non-Targeted C\&W Attack Total Net-Worth Difference ]{{\includegraphics[width=0.6\columnwidth]{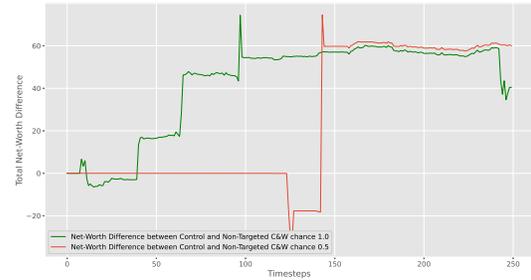} }}%
    \caption{Net-Worth Differences between Control Total Net-Worth and Non-Targeted Attacks on TensorTrade's DQN Total Net-Worth}%
    \label{tt_fgsm_nontarget_networthdiff}%
    \end{minipage}
\end{figure}

For our C\&W attack, we could not implement in the way we did for the basic DQN. C\&W optimizes over a vector $w$ that produces an adversarial example $x + \delta$ with bounded values between $[0,1]$ given$ x + \delta = \frac{1}{2}\times (tanh(w) + 1)$ with $x$ as the true sample which implies that the feature data must be normalized for a direct C\&W attack. Our agent is not trained on normalized data, but we can similarly add an amount of adversarial perturbation to the original observation like in FGSM as long as the adversarial observations falls within the distribution. This C\&W attack implementation is weaker than the basic DQN's C\&W attack implementation but still impacts the performance of the agent. We have individual scalars $k_{0} = 0.01$, $k_{1} = 1.0$, and $k_{2} = 1.0$. We consider $\epsilon = 1.0$ to be the scalar applied to each $k_d$ where $d$ is a feature/dimension. We have the learning rate set to 0.5 and max iterations of 100. Table \ref{tt_fgsm_nontarget_samples} and Table \ref{tt_cw_nontarget_samples} contain successful samples for the non-target FGSM and C\&W attack on TT's DQN. Despite manually testing scalar values, we have crafted adversarial samples that are convincing. Additionally, we have provided the total reward difference and net-worth difference between the control agent and agents under attack in Figure \ref{non_target_attacks_tt_rd} and Figure \ref{tt_fgsm_nontarget_networthdiff}, respectively. Through these results, we establish that the test-time performance of the target policy in regards to its total reward is negatively impacted by our attacks. We have also shown that the agent's net-worth is also impacted, but not necessarily reflected by total reward. 

It is noteworthy that constraints are applied after the construction of an adversarial tuple, which impact the number of failures. For example, for an FGSM attack at a timestep $t$ on the basic DQN, the gradient provides the direction of the minimum and we craft an adversarial tuple starting from the original tuple either moving away from the minimum (for non-targeted) or moving towards the minimum for a targeted class. We shift the values of the adversarial samples along the respected dimension if they violate the constraint which is not in the direction of gradient. This affects targeted attacks, but can also impact non-targeted attacks. Though not implemented, failures preserved in the history window can impact future actions while in the observation leading to a sub-optimal trajectory. Though all observations that receive positive reward from the environment increase the return, some timesteps contribute more to the return than others emphasizing that any successful attack can have varying impacts on the agent's performance for timing dependent tasks. We see that even less frequent attacks on the simpler DQN impact the agent's performance at test-time and weaker attacks like FGSM are also effective against TT's DQN agent performance.                    


\begin{table}[htbp]
\centering
\hspace*{-0.5cm}
\begin{tabular}{ccccc}                                   
\multicolumn{5}{c}{} \\\hline           
timestep & original observation & perturbed observation & a & $a^\prime$ \\
\hline
894 & 0.0,-0.00354677, -0.00354677 & 0.0000, -0.0045, -0.0025 & 1 & 0\\
3973 & 0.0, -0.00048828,-0.00048828 & 0.0000, -0.0006, -0.0004 & 1 & 0\\
9599 & 0.00294118,-0.0004902,0.00294118 &  0.0027, -0.0002,  0.0027 & 0 & 1\\
16323 & 0.00435098,0.0,         0.00290065 & 0.0041, 0.0000, 0.0032 &2 &0\\
23283 & 0.00074322,-0.00371609,0.00074322 & 0.0001, -0.0044,  0.0001 & 0 & 1\\

\hline
\end{tabular}
\caption{Successful Basic DQN Non-Target FGSM Observations Samples}
\label{fgsm_nontarget_samples}
\end{table}

\begin{table}[htbp]
\centering

\begin{tabular}{ccccc}                                   
\multicolumn{5}{c}{} \\\hline           
timestep & original observation & perturbed observation & a & $a^\prime$\\
\hline
1602 & 0.00203314, 0.0,         0.00203314 & 0.0003, 0.0000, 0.0003 & 0 & 1\\
4735 & 0.00707071, 0.0,0.00707071 & 0.0002, 0.0000, 0.0002 & 0 & 1\\
5346 & 0.0032695 ,-0.00140121,0.0032695 &  0.0002, -0.0002,  0.0002 & 0 & 1 \\
17424 &  0.0010985,-0.0010985 , 0.0010985 & 0.0002, -0.0002,  0.0002 & 2 & 0\\
29779 & 0.00039904,-0.00079808, 0.00039904 & 0.0003, -0.0003,  0.0003 & 0 & 1\\
\hline
\end{tabular}
\caption{Successful Basic DQN Non-Target C\&W Observations Samples}
\label{cw_nontarget_samples}
\end{table}

\begin{table}[htbp]
\centering
\resizebox{\columnwidth}{!}{%
\begin{tabular}{ccccc}                                   
\multicolumn{5}{c}{} \\\hline           
timestep & original observation & perturbed observation & a & $a^\prime$\\
\hline
8 &-5.4566085e-04,  1.1495910e+00 , 5.8555717e+01 &-6.254339e-03,  1.829591e+00,  5.862372e+01 & 15 (B)& 46 (S) \\
43 &1.4353440e-03, -2.1612392e+01,  2.8645555e+01 &2.2764657e-02, -2.4032393e+01,  2.8403555e+01 &96 (S) &15 (B)\\
83 &  2.9894735e-03, -6.5872850e+00,  5.6151459e+01& 1.5589474e-02, -5.3272848e+00,  5.6277458e+01& 46 (S) &153 (B)\\
97 &1.1075724e-02, 1.1280737e+01, 6.5658806e+01 &1.5242761e-03, 1.2540737e+01, 6.5784805e+01 &96(S) & 15 (B)\\
105 &-1.7102661e-04, -1.1659336e+00,  5.8726223e+01 & -3.0171026e-02, -4.1659336e+00,  5.8426224e+01& 96 (S)&15 (B)\\
\hline
\end{tabular}%
}
\caption{Successful TensorTrade Non-Target FGSM Observations Samples}
\label{tt_fgsm_nontarget_samples}
\end{table}

\begin{table}[htbp]
\centering
\resizebox{\columnwidth}{!}{%
\begin{tabular}{ccccc}                                   
\multicolumn{5}{c}{} \\\hline           
timestep & original observation & perturbed observation & a & $a^\prime$\\
\hline
83 &  2.9894735e-03, -6.5872850e+00 , 5.6151459e+01 & 1.2964869e-02, -5.5897794e+00 , 5.6153919e+01 & 46 (S) & 153 (B)\\
89 & 5.6910915e-03, -4.7228575e+00 , 5.6884933e+01& 1.5666518e-02 ,-3.7252722e+00 , 5.6887394e+01 & 179 (B)&122 (S)\\
115 & 9.5082802e-04 -5.1257310e+00  5.6966236e+01 &  1.0926254e-02 -4.1281385e+00  5.6968697e+01 & 164 (S) &15 (B) \\
219 &  1.7913189e-03, 1.8073471e-01 ,4.2434292e+01 & 1.1766739e-02, 1.1783471e+00, 4.2436752e+01 & 102 (S) &0 (W)\\
\hline
\end{tabular}
}
\caption{Successful TensorTrade Non-Target C\&W Observations Samples}
\label{tt_cw_nontarget_samples}
\end{table}

\begin{table}[htbp]
\hspace*{-0.5cm}
\centering

\begin{tabular}{cccc||cccc}
\hline\\
\multicolumn{4}{c}{FGSM} 
&                                            
\multicolumn{4}{c}{C \& W} \\\hline           
Chance& No. Attempts & No. Failures & N.C.N   & Chance& No. Attempts & No. Failures & N.C.N   \\
\hline
  0.1 & 26 & 25 &2 &0.1&17&17&0 \\
  0.5 & 123 & 117 & 7 &0.5&114&110&3 \\
  1.0 & 242 & 236 &7 &1.0&246&240&3 \\
\hline
\end{tabular}
\caption{Non-Target FGSM and C\&W Attacks Attempts and Failures on TensorTrade's DQN}
\label{tt_nontarget_table}
\end{table}

\begin{table}[htbp]
\centering
\begin{tabular}{ccc||ccc}
\hline\\
\multicolumn{3}{c}{FGSM} 
&                                            
\multicolumn{3}{c}{C \& W} \\\hline           
Chance& No. Attempts & No. Failures    & Chance& No. Attempts & No. Failures   \\
\hline
  0.01 & 286 & 6       & 0.01 & 329 & 163 \\
 0.1 & 3349 & 176       & 0.1 & 3016 & 1751  \\
  0.5 & 15818 & 3329     & 0.5 & 15979 & 9358 \\
   1.0 & 31779 & 10778       & 1.0 & 31779 & 18716 \\
\hline
\end{tabular}
\caption{Non-Target FGSM and C\&W Attacks Attempts and Failures on the Basic DQN}
\label{nontarget_table}
\end{table}


\vspace{4cm}


\subsection{Targeted Perturbation Attacks}
We also investigate targeted attacks, which aim to manipulate a policy into taking an adversarial action $a_t^{\prime}$ instead of action $a_t$ at a timestep $t$. We have evaluated against targeted FGSM and targeted C\&W attacks using $L_2$ loss for both DQNs. 

\begin{figure}%
    \centering
    \begin{minipage}{\textwidth}
        \hspace*{-2cm} 
    \subfloat[\centering Targeted FGSM Attack]{{\includegraphics[width=0.6\columnwidth]{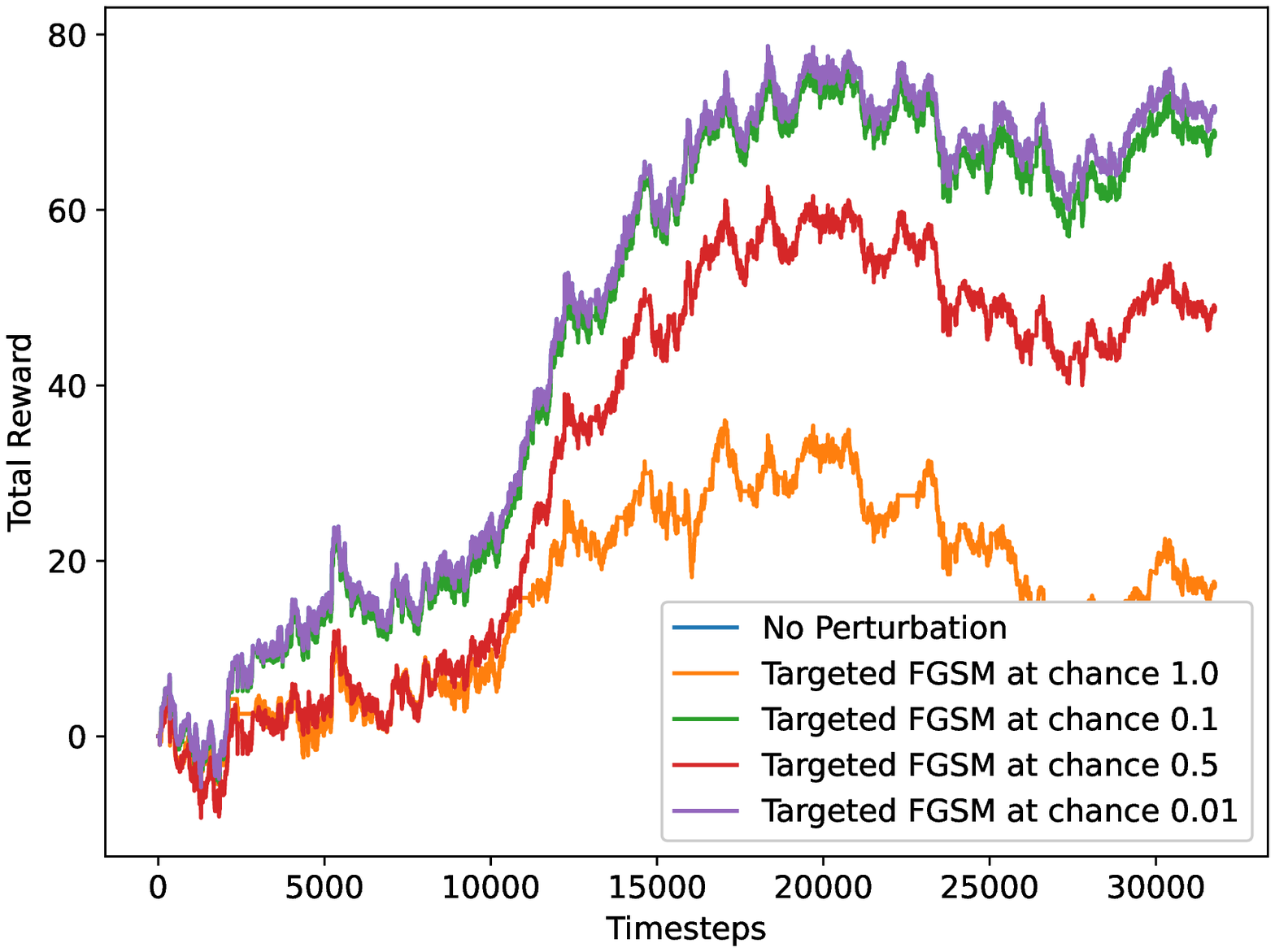}  }}%
    \qquad
    \subfloat[\centering Targeted C\&W Attack ]{{\includegraphics[width=0.6\columnwidth]{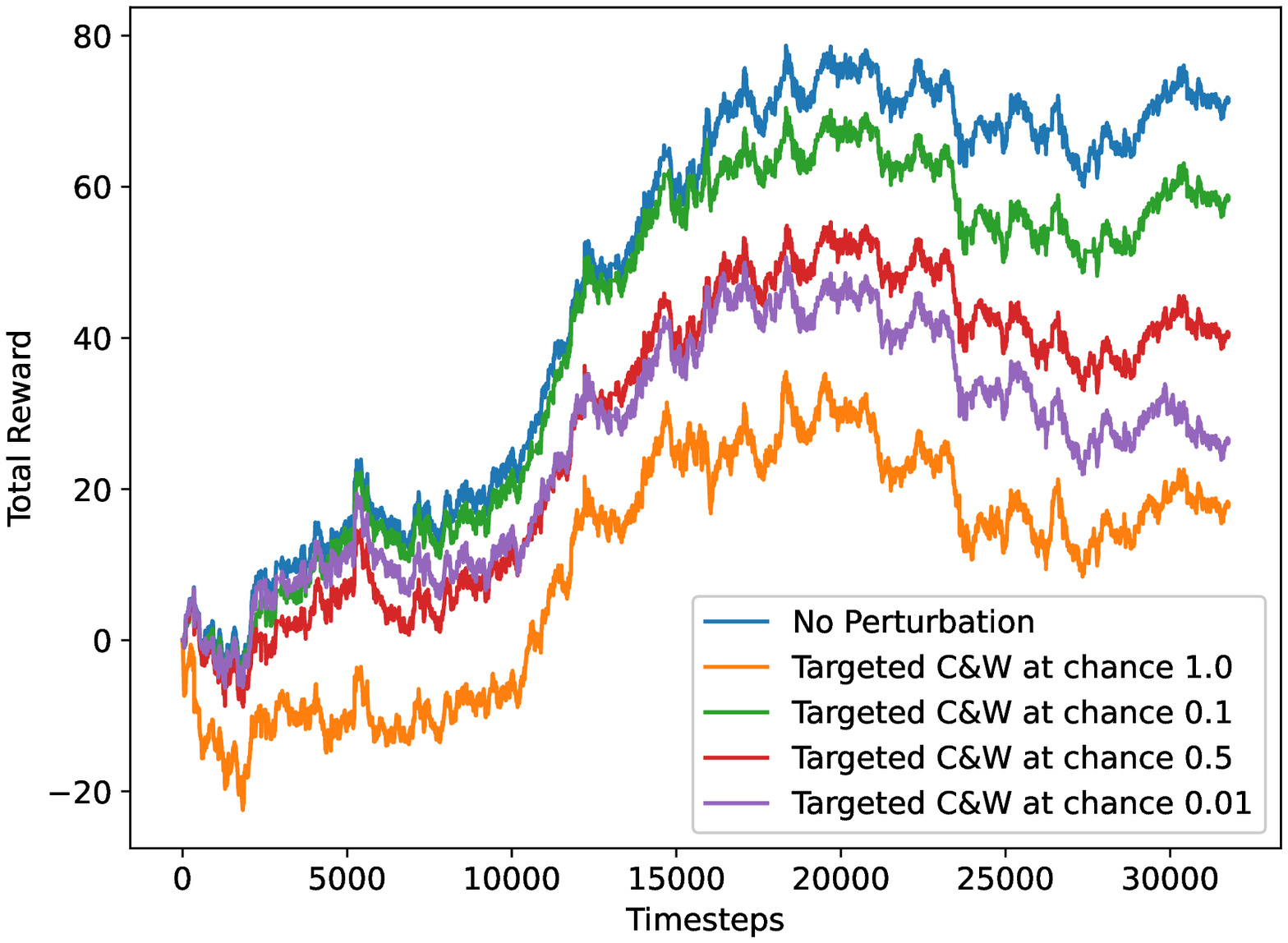}  }}%
    \caption{Targeted Attacks on Basic DQN}%
    \label{target_attacks}%
    \end{minipage}
\end{figure}

For the targeted FGSM attack on the basic DQN, we allowed up to five iterations of increasing $\epsilon$, starting at $0.0001$ where the last iteration tests at an $\epsilon$ value of 0.001. The adversarial action $a_t'$ is set to be the action with the least Q value at timestep $t$. The same constraints are implemented from the non-target attacks on the simpler DQN. C\&W parameters are the same as non-targeted C\&W attack on the simpler DQN. Table \ref{target_table} contains the failure count and attempt count. We define failure as the failure to change the agent's action $a_t$ to the adversarial action $a_t^{\prime}$ at timestep $t$. To reflect only the impact of targeted attacks, only perturbations in observation that resulted in adversarial action $a_t^{\prime}$ were kept for future timesteps. Results of the targeted FGSM and C\&W attack is outlined in Figure \ref{target_attacks} for the simpler DQN. All non-targeted attacks were assigned the optimal action $a_t$ for each respected timestep $t$. Our adversarial tuples are simple but should emphasize that adversarial attacks crafted under expensive parameters like low learning rate, high number of iterations, and high confidence can produce more human convincing adversarial samples. Successful samples for these attacks on the simpler DQN can be found in Table \ref{fgsm_target_samples} and Table \ref{cw_target_samples}.

\begin{figure}%
    \centering
    \begin{minipage}{\textwidth}
        \hspace*{-2cm} 
    \subfloat[\centering Reward Difference Targeted FGSM Attack]{{\includegraphics[width=0.6\columnwidth]{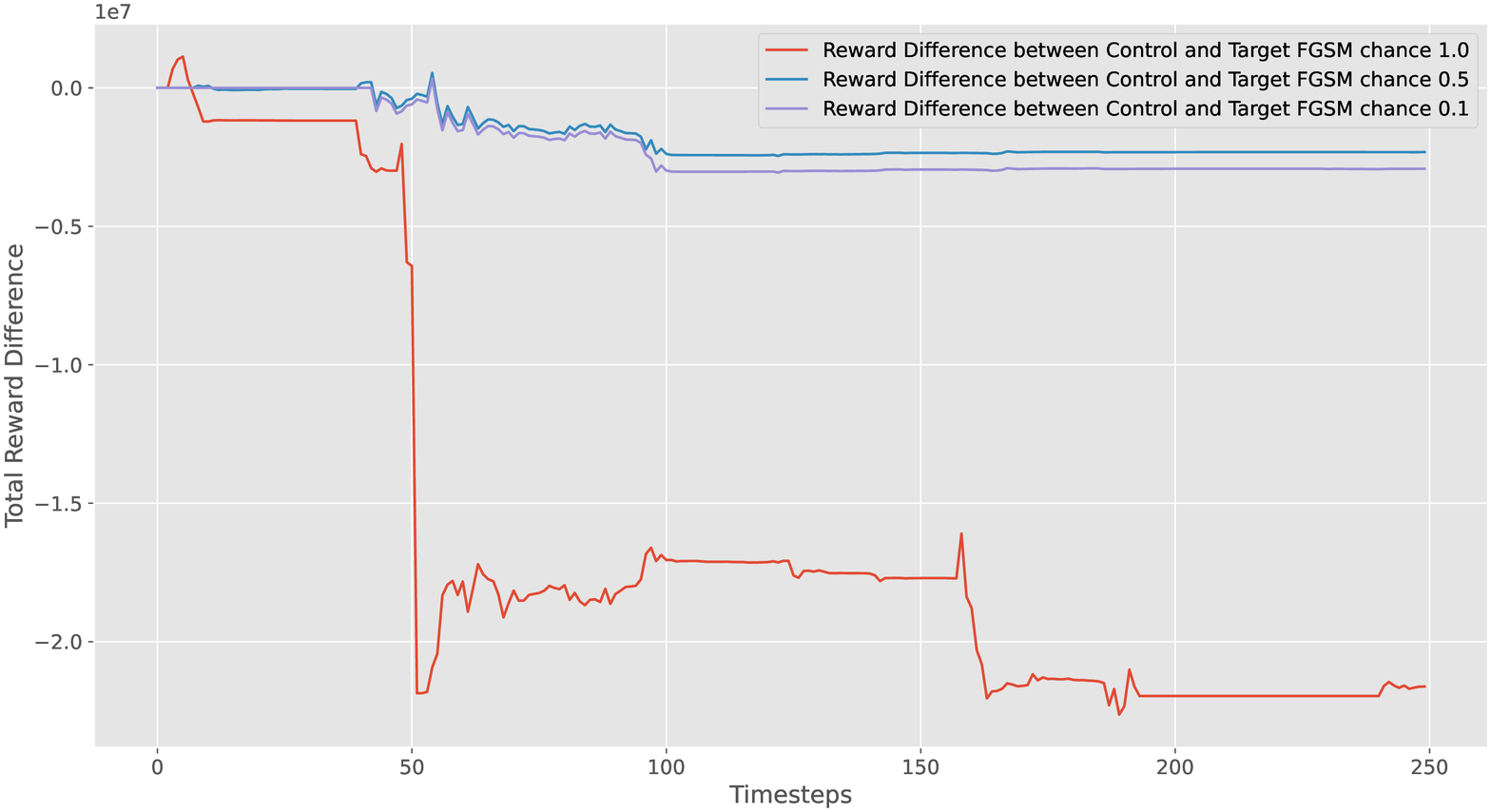}  }}%
    \qquad
    \subfloat[\centering Reward Difference Targeted C\&W Attack ]{{\includegraphics[width=0.6\columnwidth]{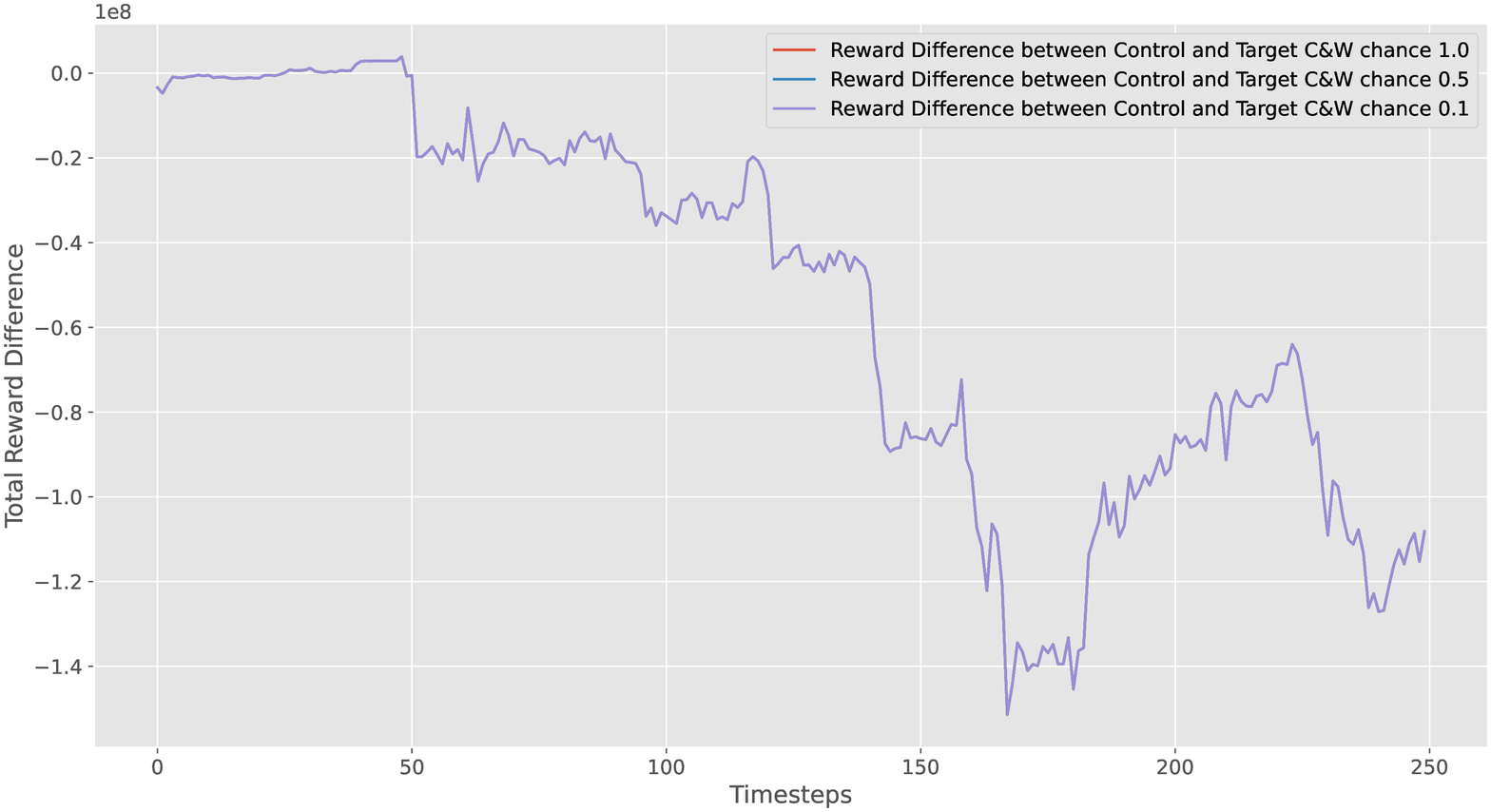}  }}%
    \caption{Reward Difference between Control Total Reward and Targeted Attacks on TensorTrade's DQN Total Reward}%
    \label{tt_target_attacks}%
    \end{minipage}
\end{figure}

\begin{figure}%
    \centering
    \begin{minipage}{\textwidth}
        \hspace*{-2cm} 
    \subfloat[\centering Net-Worth Difference Targeted FGSM Attack]{{\includegraphics[width=0.6\columnwidth]{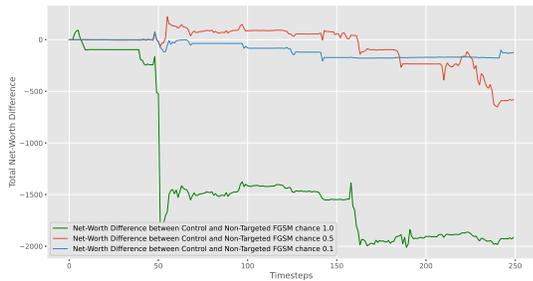}  }}%
    \qquad
    \subfloat[\centering Net-Worth Difference Targeted C\&W Attack ]{{\includegraphics[width=0.6\columnwidth]{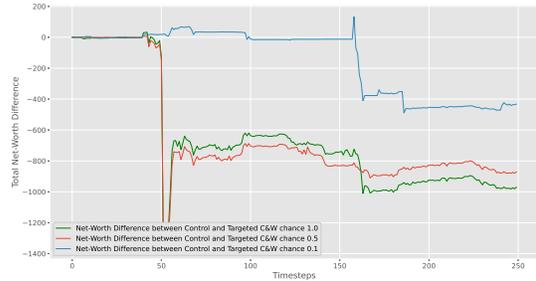}  }}%
    \caption{Net-Worth Difference between Control Total Reward and Targeted Attacks on TensorTrade's DQN Net-Worth}%
    \label{tt_target_attacks_networth}%
    \end{minipage}
\end{figure}

For the targeted FGSM attack through TT's DQN, we consider a similar setup to non-targeted FGSM attack on the simpler DQN but we will be more lenient on the failure criteria because the action space is large. We will consider an attack a failure if the attack fails to change the agent's action $a_t$ to the adversarial action $a_t^{\prime}$ at a timestep $t$. We will consider a partial success if our attack results in an action $a^m$ where the action type (buy, sell, wait) of $a^m$ is the adversarial action type for action $a_t^{\prime}$. Successful attacks and partial successful attacks will be preserved in observation, otherwise all failures reassign the agent to take optimal action $a_t$ at timestep $t$. Failure count and attempts are in Table \ref{tt_target_table_samples} and successful samples for targeted FGSM and targeted C\&W for TT's DQN can be found in Table \ref{tt_fgsm_target_samples} and Table \ref{tt_cw_target_samples} respectively. Like before for non-targeted attacks on TT's DQN, we have plots of the reward difference and net-worth difference between the control agent and attacked agents which are Figure \ref{tt_target_attacks} and \ref{tt_target_attacks_networth}. We thus establish the impact of targeted attacks on TT's DQN on test-time performance as well as its significant impact on the agent's net-worth. 

\begin{table}[h]
\centering
\begin{tabular}{ccccc}                                   
\multicolumn{5}{c}{} \\\hline           
t & x & $x^\prime$  & a & $a^\prime$ \\
\hline
301& 0.  ,  -0.0048627,-0.00243129 &0.0000, -0.0040, -0.0033 & 0 & 1\\
5254&0.00094877,-0.00332068,-0.00142315 &0.0016, -0.0027, -0.0021 & 0 & 1\\
12228 & 0.00037272,-0.00260902,-0.00111815 & 0.0012, -0.0018, -0.0018 & 2 & 1 \\
21009 & 0.0,-0.0027894, -0.00209205 & 0.0000, -0.0025, -0.0024 & 0 & 1\\
24764 &  0.00119332,-0.00357995,-0.00159109 & 0.0018, -0.0029, -0.0022 & 0 & 1 \\
\hline
\end{tabular}
\caption{Successful Basic DQN Target FGSM Observations Samples}
\label{fgsm_target_samples}
\end{table}
\begin{table}[h]
\centering
\begin{tabular}{ccccc}                                   
\multicolumn{5}{c}{} \\\hline           
t & x & $x^\prime$  & a & $a^\prime$ \\
\hline
2233&0.00490773,0.0,         0.00490773&0.0003, 0.0000, 0.0003&0&1\\
 11328&0.00041408,-0.00248447,0.00041408 & 0.0003, -0.0003,  0.0003 & 0 & 1\\
17733&0.00362319,-0.00072464, 0.00362319&0.0003, -0.0003,  0.0003&0 & 1\\
20145 & 0.00102881,0.0,0.00102881 & 0.0002, 0.0000, 0.0002 & 0 & 1\\
26787&0.00569106, 0.0,         0.00569106&0.0003, 0.0000, 0.0003&2&1\\
\hline
\end{tabular}
\caption{Successful Basic DQN Target C\&W Observations Samples }
\label{cw_target_samples}
\end{table}

\begin{table}[h]
\centering
\resizebox{\columnwidth}{!}{%
\begin{tabular}{cccccc}                                   
\multicolumn{6}{c}{} \\\hline           
t & x & $x^\prime$  & a & $a^\prime$& P.S. or S. \\
\hline
1&6.1583184e-03, 4.1991682e+00, 1.0000000e+02&3.444168e-02, 8.991680e-01, 1.009300e+02&0 (W) &144 (S) & P.S. \\
65&-4.5726676e-03,  1.6424809e+01,  6.1849476e+01&-1.4827332e-02,  2.5724810e+01,  6.2779472e+01&0 (W) & 122 (S) & P.S.\\
138&-3.4072036e-03 -4.1683779e+00  5.6641838e+01&-1.5992796e-02 -3.7883778e+00  5.7571835e+01& 0 (W)&96 (S)&P.S.\\
179&1.1997896e-06 -1.0627291e+01  6.5866417e+01&1.9401200e-02 -7.3272896e+00  6.6196411e+01& 1 (B) &78 (S) &P.S.\\
249&6.7891073e-03 -1.1369240e+01  5.6471169e+01&1.2610892e-02 -2.0669241e+01  5.5541172e+01&1 (B) &46 (S)& P.S.\\

\hline
\end{tabular}%
}
\caption{Successful TensorTrade DQN Target FGSM Observations Samples}
\label{tt_fgsm_target_samples}
\end{table}

\begin{table}[h]
\centering
\resizebox{\columnwidth}{!}{%
\begin{tabular}{cccccc}                                   
\multicolumn{6}{c}{} \\\hline           
t & x & $x^\prime$  & a & $a^\prime$ &P.S. or S. \\
\hline
1&6.1583184e-03, 4.1991682e+00, 1.0000000e+02&1.61259882e-02, 5.19593525e+00, 1.00003235e+02&0 (W) & 144 (S) & P.S. \\
2 &3.119729e-03, 8.207591e+00, 1.000000e+02&1.30873993e-02, 9.20435810e+00, 1.00003235e+02&0 (W) &15 (B) & P.S. \\
189&-5.3039943e-03 -5.4235260e+01  4.5886791e+01&-4.6636765e-03 -5.3238495e+01  4.5890026e+01&0 (W) &78 (S)& P.S.\\
244 &-5.3748242e-03,  8.7277918e+00,  5.8095055e+01&-4.5928466e-03,  9.7245588e+00,  5.8098289e+01&1 (B) &96 (S) & P.S.\\
247 &-4.9919840e-03, -9.8234949e+00,  5.4668602e+01&-4.9756868e-03, -8.8267279e+00,  5.4671837e+01&0 (W) & 15 (B) & P.S. \\

\hline
\end{tabular}%
}
\caption{Successful TensorTrade DQN Target C\&W Observations Samples }
\label{tt_cw_target_samples}
\end{table}

\begin{table}[htbp]
\makebox[\textwidth]{
\setlength\tabcolsep{1.5pt}
\begin{tabular}{cccc||cccc}

\hline\\

\multicolumn{4}{c}{FGSM} 
&                                            
\multicolumn{4}{c}{C \& W} \\\hline           
Chance& No. Attempts & No. Failures & No. Non-Target    & Chance& No. Attempts & No. Failures & No. Non-Target   \\
\hline
  0.01 & 337 & 6 & 4       & 0.01 & 327 & 294 & 89\\
 0.1 & 3148 & 191 & 98       & 0.1 & 3135 & 2915 & 903 \\
  0.5 & 15905 & 4666 & 1581     & 0.5 & 15882 & 15291 & 4837 \\
   1.0 & 31779 & 16000 & 5334      & 1.0 & 31779 & 30779 & 9953 \\
   
\hline

\end{tabular}
}
\caption{Targeted FGSM and C\&W Attacks Attempts and Failures on Basic DQN}
\label{target_table} 
\end{table}

\begin{table}[htbp]
\centering
\resizebox{\columnwidth}{!}{%
\begin{tabular}{ccccc||ccccc}
\hline\\
\multicolumn{5}{c}{FGSM} 
&                                            
\multicolumn{5}{c}{C \& W} \\\hline           
Chance& No. Attempts & No. Failures & Non-Targeted & P.S.   & Chance& No. Attempts & No. Failures & Non-Targeted & P.S.   \\
\hline
  0.1 & 248& 248&146 & 230  &0.1& 26& 26& 5 &26 \\
  0.5 & 123&123 &65 &122   &   0.5&131 &131 &25 & 127\\
  1.0 &28 &28 &9 &27   &1.0&249 &249 &70&243 \\
\hline
\end{tabular}%
}
\caption{Target FGSM and C\&W Attacks Attempts and Failures on TensorTrade's DQN}
\label{tt_target_table_samples}
\end{table}

\break
\section{Conclusion}
\label{sec:Conclusion}
We investigated the vulnerability of DRL trading agents to adversarial attacks at inference time. We identified the attack surface and vectors of algorithmic trading policies in a novel threat model, and proposed 2 attack techniques to target these policies, namely: DoS-based delay induction and MITM-based adversarial perturbation. We investigated the susceptibility of a benchmark DRL trading agent and an agent based on TensorTrade, a popular open-source framework for algorithmic trading. We implemented several adversarial attacks in non-targeted and targeted modes that follow the trading DRL threat model. Through observation perturbation of a single tuple from the history window, we demonstrated that an attacked agent can be made to perform sub-optimally. These sub-optimal trajectories result in low total reward upon test-time. Furthermore, portfolios that are tied to the agent may be impacted in ways that is not directly reflected in the performance metric at test-time, namely total reward. With TensorTrade's DQN, our attacks were shown to adversely affect the agent's net-worth. This finding may have significant repercussions on risk mitigation, as test-time performance through total reward may not alert human traders of the severity of impact upon external securities tied to the agent. 

Our experimental results also demonstrated the significant impact of inducing observational delay via DoS attacks for a single timestep. Also, we studied the resilience of DRL trading agents to perturbation attacks by implementing two whitebox adversarial example attacks. The results demonstrate that our target agents are sensitive to even weak attacks such as FGSM, as well as and more powerful attacks like C\&W. 
Furthermore, our experiments yielded that perturbing even small ratios of all observations is sufficient to incur negative impact on the agent's test-time performance. It is noteworthy that that in our experiments, adversarial examples were crafted such that they abide by the constraints of meaningful observation values and minimal perturbation, and are hence maintain some degree of plausibility to human traders. 

The reported findings establish the need for further research on various aspects of security in DRL trading agents. One such aspect is the need for metrics and measurement techniques for benchmarking the resilience and robustness of trading policies to adversarial attacks. Furthermore, our results call for further studies on mitigation and defense techniques against adversarial manipulation. These studies are likely to find current risk-aware DRL approaches of limited utility in this domain, as such techniques are typically addressing accidental (i.e., non-adversarial) noises in the dynamics of the environment. Lastly, considering the significance of R\&D efforts in developing and acquiring proprietary algorithmic trading policies, there remains a critical need to study the impact of policy imitation attacks \cite{BehzadanImitation} targeting algorithmic trading. 

\bibliographystyle{IEEEtran}
\bibliography{refs}

\end{document}